%% file: main.tex
\documentclass[10pt,twocolumn,letterpaper]{article}

\usepackage{cvpr}              %

\input{preamble}

\definecolor{cvprblue}{rgb}{0.21,0.49,0.74}
\usepackage[dvipsnames]{xcolor}
\usepackage{textcomp}
\usepackage[pagebackref,breaklinks,colorlinks,citecolor=cvprblue]{hyperref}

\def\paperID{N/A} %
\def\confName{ECCVW}
\def\confYear{2024}

\begin{document}
\title{Can Visual Foundation Models Achieve Long-term Point Tracking?}

\author{Görkay Aydemir\textsuperscript{1} \quad 
        Weidi Xie\textsuperscript{3, 4} \quad
        Fatma Güney\textsuperscript{1, 2} \\
        \textsuperscript{1} Department of Computer Engineering, Koç University \quad
        \textsuperscript{2} KUIS AI Center \\ 
        \textsuperscript{3} CMIC, Shanghai Jiao Tong University \quad
        \textsuperscript{4} Shanghai AI Laboratory \\
        \texttt{\small \{gaydemir23, fguney\}@ku.edu.tr} \quad
        \texttt{\small weidi@sjtu.edu.cn}
        }

\maketitle
\input{commands}
\input{sec/00-abstract}
\input{sec/01-intro}

\input{sec/02-rw}

\input{sec/03-methodology}

\input{sec/04-exp}
\input{sec/05-discussion}

{\small
\bibliographystyle{ieeenat_fullname}
\bibliography{bibliography_short, egbib}
}

\end{document}

%% file: preamble.tex
\usepackage[dvipsnames]{xcolor}

%% file: commands.tex
\newcommand{\Perp}{\perp\!\!\! \perp}
\newcommand{\bK}{\mathbf{K}}
\newcommand{\bX}{\mathbf{X}}
\newcommand{\bY}{\mathbf{Y}}
\newcommand{\bk}{\mathbf{k}}
\newcommand{\bx}{\mathbf{x}}
\newcommand{\by}{\mathbf{y}}
\newcommand{\bhy}{\hat{\mathbf{y}}}
\newcommand{\bty}{\tilde{\mathbf{y}}}
\newcommand{\bG}{\mathbf{G}}
\newcommand{\bI}{\mathbf{I}}
\newcommand{\bg}{\mathbf{g}}
\newcommand{\bS}{\mathbf{S}}
\newcommand{\bs}{\mathbf{s}}
\newcommand{\bM}{\mathbf{M}}
\newcommand{\bw}{\mathbf{w}}
\newcommand{\eye}{\mathbf{I}}
\newcommand{\bU}{\mathbf{U}}
\newcommand{\bV}{\mathbf{V}}
\newcommand{\bW}{\mathbf{W}}
\newcommand{\bn}{\mathbf{n}}
\newcommand{\bv}{\mathbf{v}}
\newcommand{\bwv}{\mathbf{wv}}
\newcommand{\bq}{\mathbf{q}}
\newcommand{\bR}{\mathbf{R}}
\newcommand{\bi}{\mathbf{i}}
\newcommand{\bj}{\mathbf{j}}
\newcommand{\bp}{\mathbf{p}}
\newcommand{\bt}{\mathbf{t}}
\newcommand{\bJ}{\mathbf{J}}
\newcommand{\bu}{\mathbf{u}}
\newcommand{\bB}{\mathbf{B}}
\newcommand{\bD}{\mathbf{D}}
\newcommand{\bz}{\mathbf{z}}
\newcommand{\bP}{\mathbf{P}}
\newcommand{\bC}{\mathbf{C}}
\newcommand{\bA}{\mathbf{A}}
\newcommand{\bZ}{\mathbf{Z}}
\newcommand{\bff}{\mathbf{f}}
\newcommand{\bF}{\mathbf{F}}
\newcommand{\bo}{\mathbf{o}}
\newcommand{\bO}{\mathbf{O}}
\newcommand{\bc}{\mathbf{c}}
\newcommand{\bm}{\mathbf{m}}
\newcommand{\bT}{\mathbf{T}}
\newcommand{\bQ}{\mathbf{Q}}
\newcommand{\bL}{\mathbf{L}}
\newcommand{\bl}{\mathbf{l}}
\newcommand{\ba}{\mathbf{a}}
\newcommand{\bE}{\mathbf{E}}
\newcommand{\bH}{\mathbf{H}}
\newcommand{\bd}{\mathbf{d}}
\newcommand{\br}{\mathbf{r}}
\newcommand{\be}{\mathbf{e}}
\newcommand{\bb}{\mathbf{b}}
\newcommand{\bh}{\mathbf{h}}
\newcommand{\bhh}{\hat{\mathbf{h}}}
\newcommand{\btheta}{\boldsymbol{\theta}}
\newcommand{\bTheta}{\boldsymbol{\Theta}}
\newcommand{\bpi}{\boldsymbol{\pi}}
\newcommand{\bphi}{\boldsymbol{\phi}}
\newcommand{\bpsi}{\boldsymbol{\psi}}
\newcommand{\bPhi}{\boldsymbol{\Phi}}
\newcommand{\bmu}{\boldsymbol{\mu}}
\newcommand{\bsigma}{\boldsymbol{\sigma}}
\newcommand{\bSigma}{\boldsymbol{\Sigma}}
\newcommand{\bGamma}{\boldsymbol{\Gamma}}
\newcommand{\bbeta}{\boldsymbol{\beta}}
\newcommand{\bomega}{\boldsymbol{\omega}}
\newcommand{\blambda}{\boldsymbol{\lambda}}
\newcommand{\bLambda}{\boldsymbol{\Lambda}}
\newcommand{\bkappa}{\boldsymbol{\kappa}}
\newcommand{\btau}{\boldsymbol{\tau}}
\newcommand{\balpha}{\boldsymbol{\alpha}}
\newcommand{\nR}{\mathbb{R}}
\newcommand{\nN}{\mathbb{N}}
\newcommand{\nL}{\mathbb{L}}
\newcommand{\cN}{\mathcal{N}}
\newcommand{\cA}{\mathcal{A}}
\newcommand{\cM}{\mathcal{M}}
\newcommand{\cR}{\mathcal{R}}
\newcommand{\cB}{\mathcal{B}}
\newcommand{\cG}{\mathcal{G}}
\newcommand{\cL}{\mathcal{L}}
\newcommand{\cH}{\mathcal{H}}
\newcommand{\cS}{\mathcal{S}}
\newcommand{\cT}{\mathcal{T}}
\newcommand{\cO}{\mathcal{O}}
\newcommand{\cC}{\mathcal{C}}
\newcommand{\cP}{\mathcal{P}}
\newcommand{\cE}{\mathcal{E}}
\newcommand{\cI}{\mathcal{I}}
\newcommand{\cF}{\mathcal{F}}
\newcommand{\cK}{\mathcal{K}}
\newcommand{\cV}{\mathcal{V}}
\newcommand{\cY}{\mathcal{Y}}
\newcommand{\cX}{\mathcal{X}}
\newcommand{\cZ}{\mathcal{Z}}
\def\bgamma{\boldsymbol\gamma}

 \newcommand{\mytexttilde}{\raisebox{0.5ex}{\texttildelow}}

\newcommand{\specialcell}[2][c]{%
  \begin{tabular}[#1]{@{}c@{}}#2\end{tabular}}

\newcommand{\figref}[1]{\Fig~\ref{#1}}
\newcommand{\secref}[1]{Section~\ref{#1}}
\newcommand{\algref}[1]{Algorithm~\ref{#1}}
\newcommand{\eqnref}[1]{Eq.~\ref{#1}}
\newcommand{\tabref}[1]{Table~\ref{#1}}

\newcommand{\rulesep}{\unskip\ \vrule\ }

\renewcommand{\b}{\ensuremath{\mathbf}}

\def\mc{\mathcal}
\def\mb{\mathbf}

\newcommand{\T}{^{\raisemath{-1pt}{\mathsf{T}}}}

\makeatletter
\DeclareRobustCommand\onedot{\futurelet\@let@token\@onedot}
\def\@onedot{\ifx\@let@token.\else.\null\fi\xspace}
\def\eg{{\em e.g}\onedot} \def\Eg{E.g\onedot}
\def\ie{{\em i.e}\onedot} \def\Ie{I.e\onedot}
\def\cf{cf\onedot} \def\Cf{Cf\onedot}
\def\etc{{\em etc}\onedot} \def\vs{vs\onedot}
\def\wrt{wrt\onedot}
\def\dof{d.o.f\onedot}
\def\etal{et~al\onedot} \def\iid{i.i.d\onedot}
\def\Fig{Fig\onedot} \def\Eqn{Eqn\onedot} \def\Sec{Sec\onedot} \def\Alg{Alg\onedot}
\makeatother

\newcommand{\xdownarrow}[1]{%
  {\left\downarrow\vbox to #1{}\right.\kern-\nulldelimiterspace}
}

\newcommand{\xuparrow}[1]{%
  {\left\uparrow\vbox to #1{}\right.\kern-\nulldelimiterspace}
}

\renewcommand\UrlFont{\color{magenta}\rmfamily}

\newcommand*\rot{\rotatebox{90}}
\newcommand{\boldparagraph}[1]{\noindent{\bf #1:} }
\newcommand{\boldquestion}[1]{\noindent{\bf #1} }

\newcommand{\ftm}[1]{ \noindent {\color{cyan} {\bf Fatma:} {#1}} }
\newcommand{\ga}[1]{ \noindent {\color{blue} {\bf Gorkay:}  {#1}} }
\newcommand{\mrb}[1]{ \noindent {\color{magenta} {\bf Weidi:}  {#1}} }

\newcommand{\cmark}{\ding{51}}%
\newcommand{\xmark}{\ding{55}}%

%% file: sec/00-abstract.tex
\begin{abstract}
Large-scale vision foundation models have demonstrated remarkable success across various tasks, underscoring their robust generalization capabilities. While their proficiency in two-view correspondence has been explored, their effectiveness in long-term correspondence within complex environments remains unexplored. To address this, we evaluate the geometric awareness of visual foundation models in the context of point tracking: (i) in zero-shot settings, without any training; (ii) by probing with low-capacity layers; (iii) by fine-tuning with Low Rank Adaptation (LoRA). Our findings indicate that features from Stable Diffusion and DINOv2 exhibit superior geometric correspondence abilities in zero-shot settings. Furthermore, DINOv2 achieves performance comparable to supervised models in adaptation settings, demonstrating its potential as a strong initialization for correspondence learning.

\end{abstract}

%% file: sec/01-intro.tex
\section{Introduction}

\quad Large-scale vision foundation models have achieved significant success across various tasks, including zero-shot applications like segmentation~\cite{Wang2023CVPR}, classification~\cite{Radford2021ICML}, and localization~\cite{Melas2022CVPR}, demonstrating strong generalization capabilities. Specifically, several works have explored the correspondence abilities of them~\cite{Tang2023NeurIPS, Zhang2024CVPR, El2024CVPR}, as image correspondence is a fundamental problem in computer vision, crucial for tasks such as 3D reconstruction~\cite{Schonberger2016CVPR}, image editing~\cite{Mou2024ICLR}, and style transfer~\cite{Lee2020CVPR}. However, this exploration has been limited to the two-view correspondence.

Point tracking extends the correspondence estimation problem into a long-term context~\cite{Harley2022ECCV, Doersch2022NeurIPS}. The objective of this task is to estimate the 2D projection, \ie the location in the video frame, of the same physical point throughout the video. This involves overcoming variations in appearance, visual changes, and occlusions. Evaluating foundation models' performance in point tracking is therefore crucial for understanding their geometric awareness in complex scenarios. This task requires a high level of geometric understanding to consistently track the same physical point over time, surpassing the challenges of simple two-view correspondence. Additionally, accurate point tracking is vital for various applications, including object tracking~\cite{Rajivc2023ARXIV}, robotics~\cite{Vecerik2023ICRA}, structure-from-motion~\cite{Cang2023ARXIV}, and visual odometry~\cite{Chen2024CVPR}. Leveraging the generalization capabilities of foundation models can significantly enhance their effectiveness in these areas.

In this paper, we aim to evaluate the geometric awareness of visual foundation models for the long-term point tracking task under three settings: (i) Zero-Shot: Here, we use the frozen model directly. An ideal geometry-aware model should extract similar features for the projected pixel of the same 3D point under different views~\cite{Choy2016NeurIPS}, thereby enabling accurate point tracking over time by simply selecting the most similar locations. (ii) Probing: In this setting, we train low-capacity projections on top of the frozen model to probe the geometric awareness encoded within it~\cite{Zhan2023ARXIV}. (iii) Adaptation: We use Low Rank Adaptation (LoRA) to fine-tune the model, a technique used to adapt large language models to specific tasks~\cite{Hu2022ICLR}. This exploration will help determine if these models serve as a robust initialization under constrained training settings, with limited number of learnable parameters and modest resources. %

Our investigations reveal the following: (i) Stable Diffusion exhibits superior tracking performance among other models, followed by DINOv2. This indicates that Stable Diffusion has a better understanding of geometric correspondence, consistent with previous work~\cite{Zhan2023ARXIV, Tang2023NeurIPS}. (ii) DINOv2 can match the performance of supervised models~\cite{Doersch2022NeurIPS} with a lighter training setup, suggesting that DINOv2 serves as a strong initialization.

%% file: sec/02-rw.tex
\section{Related Work}
\label{sec:rw}

\boldparagraph{Vision Foundation Models} 
In recent years, the availability of large-scale datasets and increased computing power has led to the development of deep learning models that can handle various visual tasks. These models are trained on large amounts of data through methods like generative objectives~\cite{Rombach2022CVPR}, self-supervision~\cite{Caron2021ICCV, Oquab2023ARXIV}, or supervised learning on extensive labeled datasets~\cite{Kirillov2023ICCV}. The flexibility of these models allows them to perform different  tasks, such as video object segmentation~\cite{Wang2023CVPR, Wang2023ARXIV}, estimating correspondence~\cite{Hedlin2024NeurIPS}, object-centric learning~\cite{Aydemir2024NeurIPS}, discovering parts~\cite{Amir2021ARXIV}, and open-vocabulary segmentation~\cite{Xu2023CVPR}. \\

\boldparagraph{Point Tracking} 
Tracking points across video sequences, also known as point tracking, as defined and benchmarked by TAP-Vid~\cite{Doersch2022NeurIPS}, presents substantial challenges due to the necessity of continuously following selected points over an entire video, especially in long-term scenarios and in the presence of occlusions. PIPs~\cite{Harley2022ECCV} revisits this concept by predicting motion through iterative updates within temporal windows. TAPIR~\cite{Doersch2023ICCV} improved upon PIPs by introducing better initialization and enhanced estimations using depthwise convolutions over time. More recently, BootsTAPIR~\cite{Doersch2024ARXIV} boosted TAPIR's performance by refining it on a large number of real-world videos in a self-supervised manner. Additionally, CoTracker~\cite{Karaev2024ECCV} addresses the problem by jointly tracking multiple points and leveraging spatial correlations between them.

%% file: sec/03-methodology.tex
\section{Methodology}
\label{sec:method}

\subsection{Models}
\label{sec:method:models}
\quad We consider a range of visual models, each trained with different objectives on diverse datasets, inspired by previous work~\cite{El2024CVPR}. Specifically, we evaluate models trained with self-supervision: Masked Autoencoders (MAE)~\cite{He2022CVPR}, DINO~\cite{Caron2021ICCV}, DINOv2~\cite{Oquab2023ARXIV}, and DINOv2 with registers (DINOv2-Reg)~\cite{Darcet2023ARXIV}; language supervision: CLIP~\cite{Radford2021ICML}; image generation: Stable Diffusion (SD)~\cite{Rombach2022CVPR}; and direct supervision for classification and segmentation respectively: DeIT III~\cite{Touvron2022ECCV} and Segment Anything (SAM)~\cite{Kirillov2023ICCV}. For each model, we use the available pre-trained checkpoints.

All models, except SD, utilize Vision Transformer (ViT) architectures, while SD employs a U-Net based model. For the ViTs, we use the Base architecture (ViT-B) with a patch size of either 14 (for the DINO family) or 16 (for the remaining ViTs), unless explicitly stated otherwise. We use the output of the last block as our feature map. %

For SD, we follow the approach outlined in DIFT~\cite{Tang2023NeurIPS} and use the outputs from an intermediate layer of the U-Net encoder. Specifically, we utilize the outputs of the upsample block at $n = 2$, after noising the input frame with a time step of $t = 51$, corresponding to $1/8^{\text{th}}$ of the input image resolution. To reduce the stochasticity of the noising process, we calculate the average feature over 8 runs, each with different noisy versions of the same input image.

\subsection{Foundation Models for Point Tracking}
\label{sec:method:fomos}
\quad In this section, we explain how we leverage foundation models to estimate correspondence. For all settings, we utilize correlation maps to achieve this.

\begin{figure}
    \centering
    \includegraphics[width=.99\linewidth, trim={1.25cm 0.75cm 1cm 0.75cm}, clip]{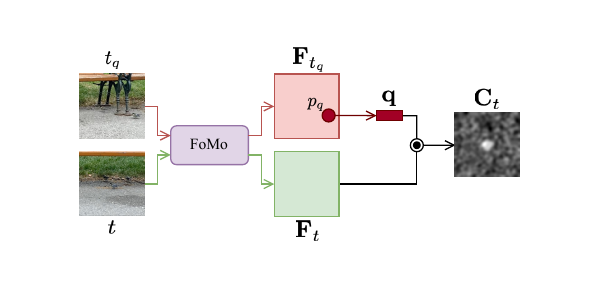}
    \caption{\textbf{Correlation Map.} The correlation map represents the similarity between frame features $\bF_{t}$ and a query feature $\bq$. %
    }
    \label{fig:cost_volume}
\end{figure}

Formally, given an encoded video, \ie feature maps, represented by a selected foundation model, $\bF \in \mathbb{R}^{T \times D \times H \times W}$, and a query prompted at frame $t_q$ at location $\bp_q$, we extract the query feature $\mathbf{q} = \bF_{t_q}(\bp_q) \in \mathbb{R}^{D \times 1 \times 1}$ by bilinear interpolation. The correlation map at any time $t$, $\bC_t \in \nR ^{H \times W}$, is then calculated using the cosine similarity between $\bF_t$ and $\mathbf{q}$~(see~\figref{fig:cost_volume}):
\begin{equation}
    \bC_t = \dfrac{\bq~.~\bF_t}{\| \bq \|~.~\| \bF_t \|} 
\end{equation}

As different models have varying stride values (8 for SD, 14 for the DINO family, and 16 for others), we ensure a fair evaluation by maintaining the same size of the correlation map, \ie the resolution of the final feature map, across different models. This is achieved by resizing the input images so that the feature maps have a consistent $32 \times 32$ resolution. \\

\boldparagraph{Zero-Shot Evaluation} In the zero-shot setting, we directly consider the most similar feature location to the query as the prediction for the corresponding time step, $t$. In other words, the predicted location $\hat{\bp}_t$ is the index of the highest value in $\bC_t$:
\begin{equation}
    \hat{\bp}_t = \underset{\bp}{\mathrm{argmax}}\, \bC_t(\bp) 
\end{equation}
However, this formulation does not provide information about the visibility of the point and is valid only if the point is visible. Therefore, we evaluate only the visible points. \\

\boldparagraph{Probing}
In TAPNet~\cite{Doersch2022NeurIPS}, correlation maps are computed using the outputs of a learnable CNN. These correlation maps are then encoded into occlusion logits and predicted locations through a series of low-capacity convolutional layers. The backbone CNN is specifically trained to develop geometric awareness, as the input to the final layers is merely a similarity map. TAPNet's low-capacity layers serve as an effective setup to assess the geometric awareness of frozen models for probing, as these layers simply project the correlation maps to outputs, with the exact correspondence information derived from the backbone.

Building on the approach of TAPNet, we employ low-capacity convolutional branches with a total of 5.5K learnable parameters. The input to these branches is the correlation map, which is calculated using features extracted from a frozen foundation model. Formally, the point prediction $\hat{\bp}_t$ and occlusion probability $\hat{\bo}_t$ are calculated as follows:
\begin{equation}
    \begin{aligned}
        \hat{\bp}_t &= \text{soft-argmax}(\Phi_{p} \circ \Phi_{e}(\bC_t)) \\
        \hat{\bo}_t &= \sigma(\Phi_{o} \circ \Phi_{e}(\bC_t))
    \end{aligned}
\end{equation}
Here,  $\Phi_{p}$ represents the point module, $\Phi_{o}$ represents the occlusion module, $\Phi_{e}$ is the common convolutional layer for both branches, $\sigma$ is the sigmoid function, and $\circ$ is function composition. The loss functions used are binary cross-entropy for occlusion and Huber loss for point prediction. For more details about the modules, please refer to the TAPNet paper. \\

\boldparagraph{Adaptation} In the adaptation phase, we go beyond mere probing and update the model to examine whether these models provide a good starting point for learning correspondence in a constrained learning setup. Rather than fine-tuning the entire model, we employ Low Rank Adaptation (LoRA)~\cite{Hu2022ICLR}. This approach maintains the integrity of the pretrained model's information while enabling adaptation with significantly fewer parameters compared to full model fine-tuning. Specifically, we apply LoRA to the query and value projections within the attention layers of the Vision Transformer (ViT), following the methodology in MeLo~\cite{Zhu2023ARXIV}. For details on the weight updates using LoRA, please refer to the MeLo paper.

%% file: sec/04-exp.tex
\section{Experiments}
\label{sec:exp}

\subsection{Experimental Setup}
\boldparagraph{Datasets} We evaluate the models on the TAP-Vid benchmark, utilizing three datasets: (i) \textbf{TAPVid-DAVIS} includes 30 real-world videos, each with around 100 frames, featuring intricate motions; (ii) \textbf{TAPVid-RGB-Stacking} is a synthetic dataset containing 50 videos of robotic manipulation tasks; (iii) \textbf{TAPVid-Kinetics} consists of over 1000 online videos with various actions. For training in probing and adaptation, we use \textbf{TAPVid-Kubric}, a synthetic simulation dataset of 11K videos, each with a fixed length of 24 frames.
\input{tables/zero_shot}

\input{fig/main_qual}

\boldparagraph{Metrics} Consistent with prior work~\cite{Doersch2023ICCV, Doersch2024ARXIV, Karaev2024ECCV}, we evaluate tracking performance using multiple metrics. Occlusion Accuracy (OA) evaluates the correctness of occlusion predictions. $\delta^\text{vis}_\text{avg}$ indicates the proportion of visible points tracked within 1, 2, 4, 8, and 16 pixels, averaged across them. Average Jaccard (AJ) combines both occlusion and prediction accuracy. Evaluations are performed in a queried-first mode, where the first visible point for each trajectory serves as the query.

\subsection{Zero-Shot Evaluation}
\boldparagraph{Different Models} The results of different foundation models in the zero-shot setting are presented in \tabref{tab:zero_shot}. We also report the performance of a supervised model, TAPNet~\cite{Doersch2022NeurIPS}, as it is the only supervised model with a final feature map resolution of $32 \times 32$. The DINO family demonstrates superior performance on DAVIS. Conversely, SD and SAM outperform the DINO models on RGB-Stacking. One possible reason for this is that RGB-Stacking is a synthetic dataset, giving SD an advantage due to its more diverse training set, which includes both real-world and synthetic data. Additionally, SD performs the best on Kinetics. On average, SD achieves the highest performance, followed by DINOv2. We visualize correlation maps of SD, DINOv2, and SAM in \figref{fig:qualitative}. SD appears to have greater geometric sensitivity, while DINOv2 demonstrates higher semantic capability. Overall, the superior performance of these two models aligns with prior work~\cite{El2024CVPR, Zhan2023ARXIV}, which shows that DINOv2 and SD have a better 3D understanding compared to other models. \\

\input{tables/zero_shot_architecture}

\boldparagraph{Effect of Architectural Variations} 
We conducted an ablation study on: (i) architectural differences to understand how the backbone capacity of the same pre-training method affects correspondence awareness, and (ii) input resolution to examine the effect of resolution, a key factor in correspondence~\cite{Doersch2023ICCV}, in \tabref{tab:zero_shot_arch} using DINOv2. The results indicate that using larger architectures consistently improves performance, with an increase from 37.1 (ViT-S) to 40.0 (ViT-G). Additionally, the resolution of the final feature map has a significant impact. Higher resolutions lead to a performance gain of 10.5 ($64 \times 64$), while lower resolutions result in a reduction of 13.1 ($16 \times 16$).

\input{tables/probing}

\subsection{Probing and Adaptation}
\quad We chose DINOv2 for probing and adaptation due to its top-2 performance in zero-shot evaluation and its resource efficiency compared to SD. For computational efficiency, we selected the ViT-S/14 architecture. For both settings, we used AdamW~\cite{Loshchilov2019ICLR}, a batch size of 16 (1/4 of TAPNet), a learning rate of $1 \times 10^{-3}$, linear warm-up, and cosine decay.  We trained for 20 epochs (\mytexttilde 1/4 iterations of TAPNet) for probing and 40 epochs (\mytexttilde 1/2 iterations of TAPNet) for adaptation. The applied weight decays were $1 \times 10^{-3}$ for probing and $1 \times 10^{-5}$ for adaptation.

We compared probing and adaptation results with different ranks, as shown in~\tabref{tab:probing}. By only probing correlation maps, DINOv2 can surpass TAPNet in OA. Moreover, adaptation of any rank performs better than TAPNet across all metrics. Surpassing supervised models even in a significantly constrained training setup, where the number of learnable parameters is substantially lower (2.5\% for rank 16), demonstrates that DINOv2 could serve as a strong initialization for correspondence learning.

%% file: tables/zero_shot.tex
\begin{table}[t]
    \centering
    \small
    \caption{\textbf{Zero-Shot Evaluation.} These results show the zero-shot evaluation results on the TAP-Vid datasets. $\delta^\text{vis}_\text{avg}$ is reported.}
    \begin{tabular}{l | c c c | r }
        \toprule
        \textbf{Model} & DAVIS & RGB-St. & Kinetics & \textbf{Avg.}\\
        \midrule
        MAE~\cite{He2022CVPR} & 23.5 & 43.2 & 27.6 & 31.4 \\
        DeIT III~\cite{Touvron2022ECCV} & 24.0 & 23.3 & 22.4 & 23.2 \\
        CLIP~\cite{Radford2021ICML} & 25.4 & 33.8 & 25.0 & 28.1 \\
        SAM~\cite{Kirillov2023ICCV} & 29.5 & \underline{44.7} & 31.4 & 35.2\\
        SD~\cite{Rombach2022CVPR} & 33.9 & \textbf{46.2} & \textbf{37.2} & \textbf{39.1} \\
        DINO~\cite{Caron2021ICCV} & 34.5 & 39.3 & 34.4 & 36.1 \\
        DINOv2-Reg~\cite{Darcet2023ARXIV} & \underline{37.4} & 35.9 & 33.6 & 35.6 \\
        DINOv2~\cite{Oquab2023ARXIV} & \textbf{38.0} & 37.8 & \underline{34.5} & \underline{36.8}\\
        \midrule
        \color{gray} TAPNet~\cite{Doersch2022NeurIPS} & \color{gray} 48.6 & \color{gray} 68.1 & \color{gray} 54.4 & \color{gray} 57.0 \\
        \bottomrule
    \end{tabular}
    \label{tab:zero_shot}
\end{table}

%% file: fig/main_qual.tex
\begin{figure*}[!h]
    \centering
    \begin{minipage}{0.33\textwidth}
        \subfloat{\includegraphics[width=.99\linewidth, trim={0cm 1.5cm 0cm 2cm}, clip]{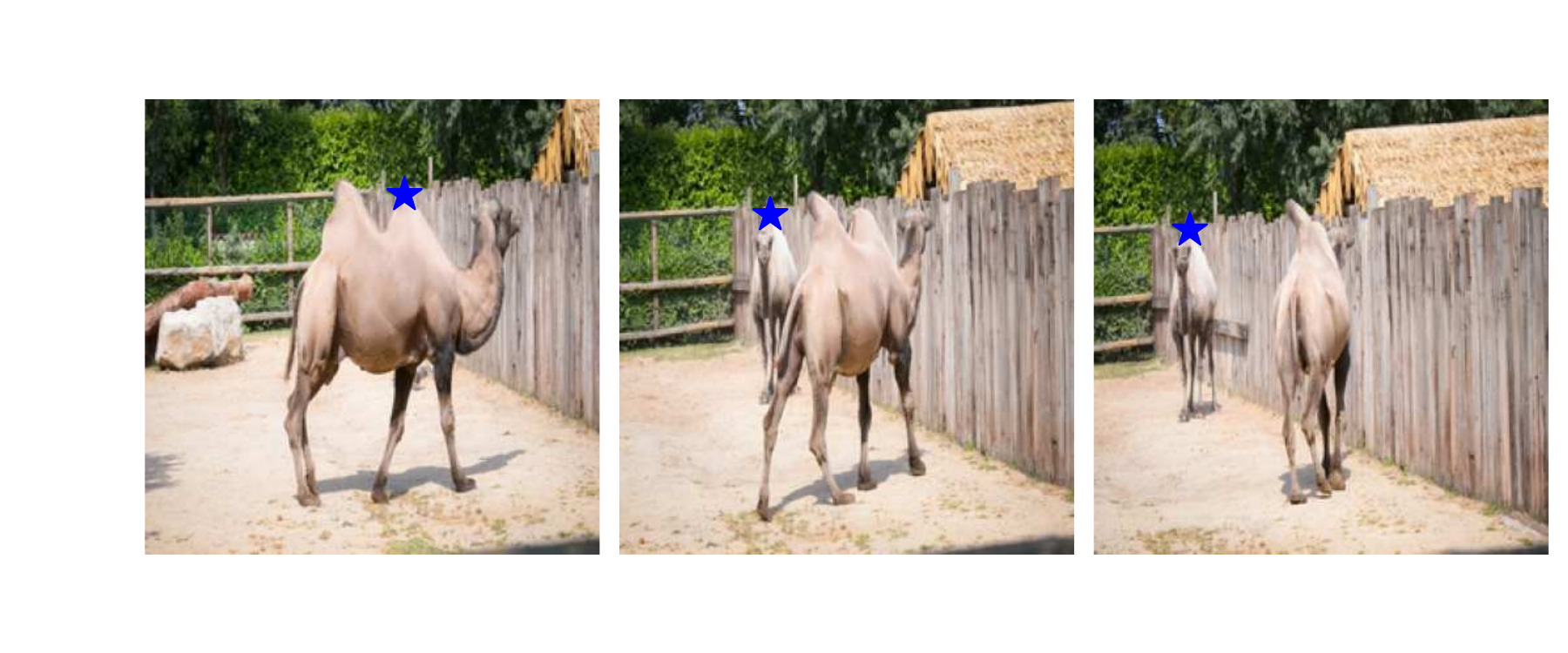}}\vspace{-0.1cm}
        \subfloat{\includegraphics[width=.99\linewidth, trim={0cm 1.5cm 0cm 2cm}, clip]{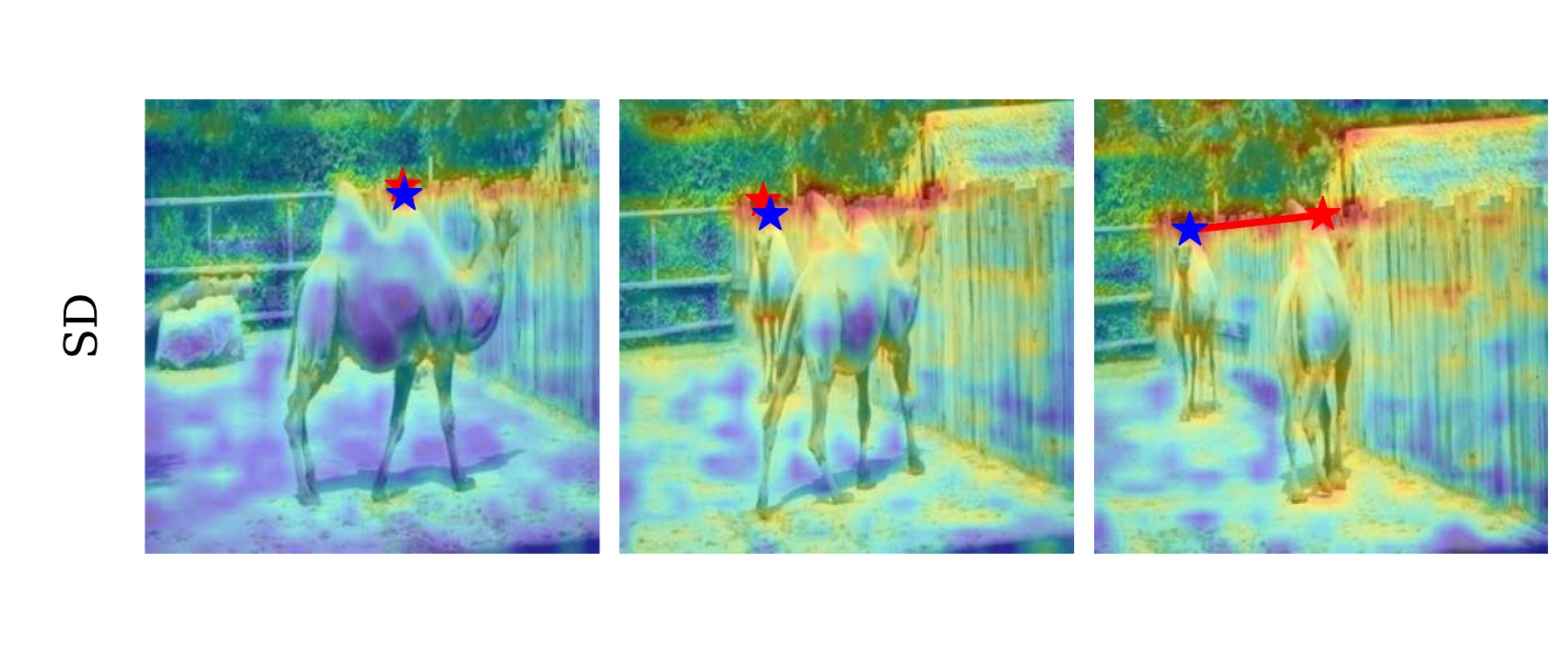}}\vspace{-0.1cm}
        \subfloat{\includegraphics[width=.99\linewidth, trim={0cm 1.5cm 0cm 2cm}, clip]{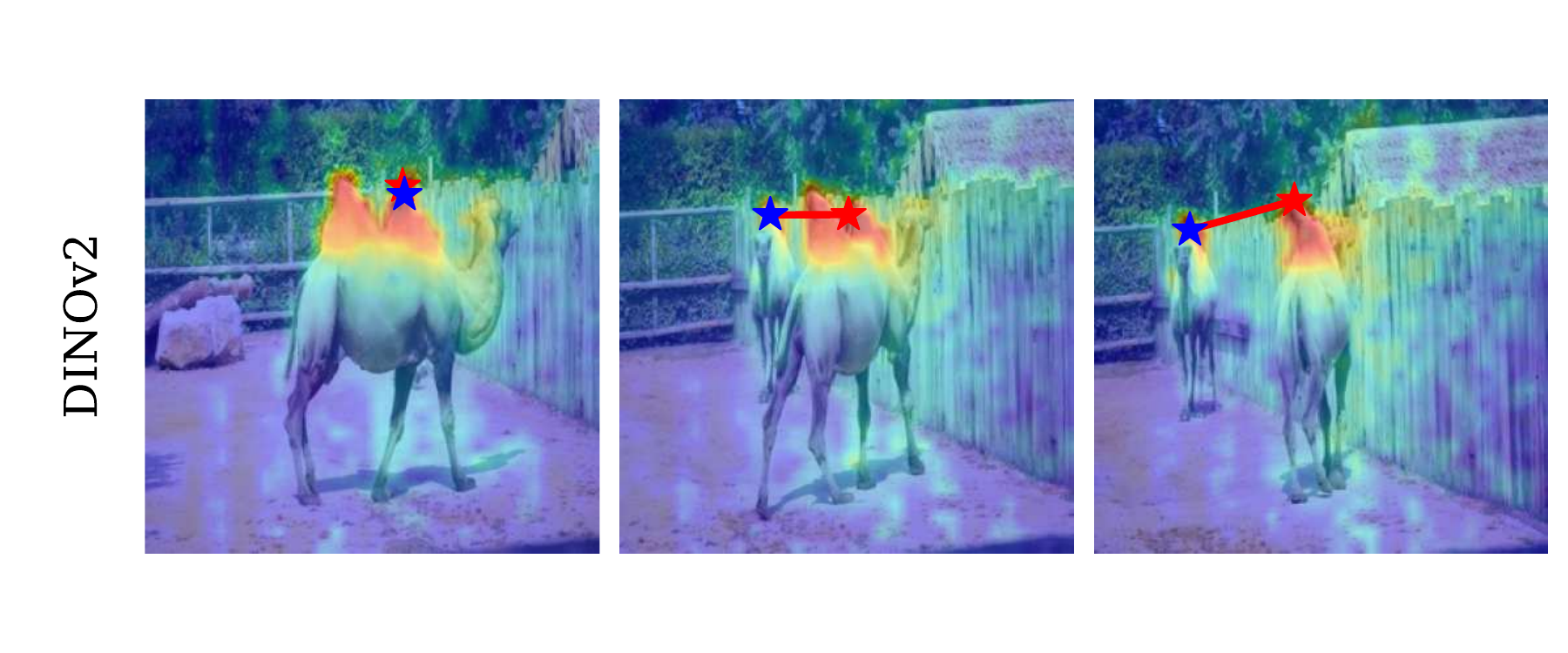}}  \vspace{-0.1cm}
        \subfloat{\includegraphics[width=.99\linewidth, trim={0cm 1.5cm 0cm 2cm}, clip]{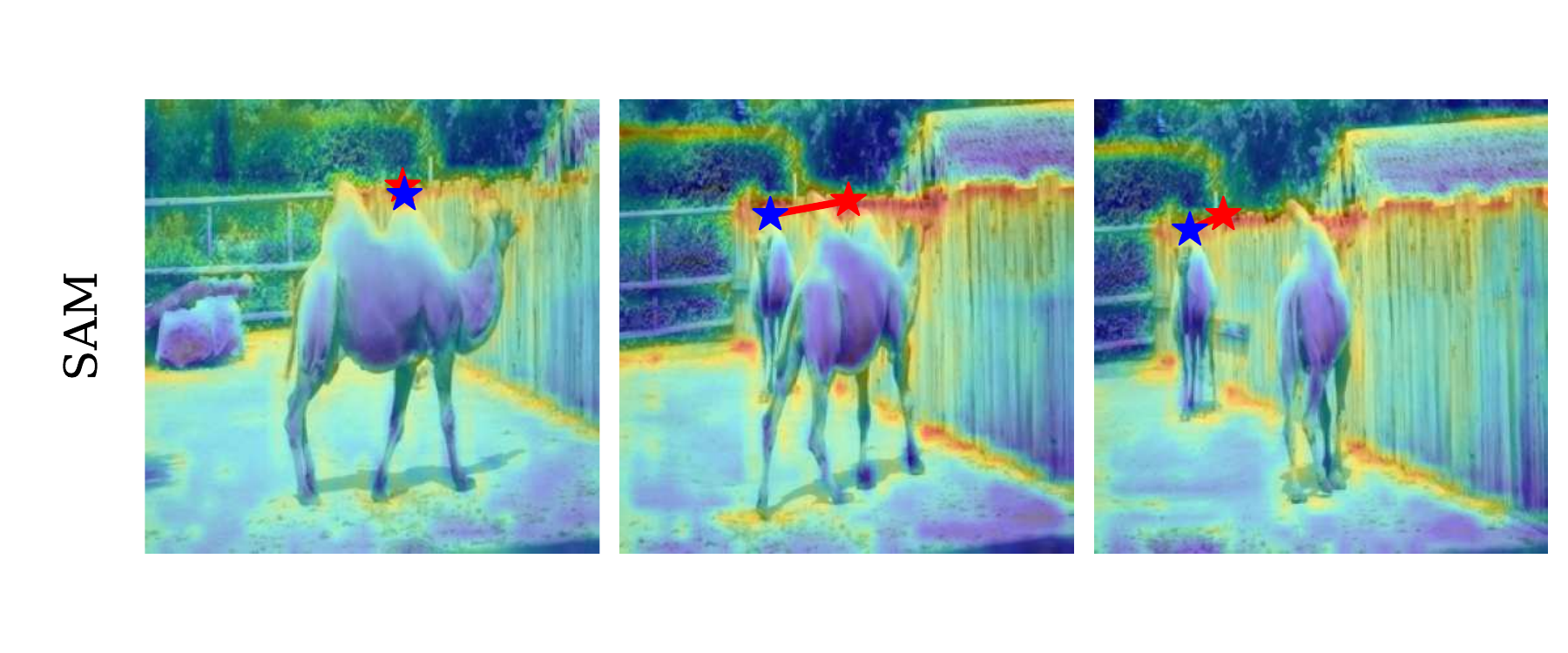}}  
    \end{minipage}%
    \begin{minipage}{0.33\textwidth}
        \subfloat{\includegraphics[width=.99\linewidth, trim={0cm 1.5cm 0cm 2cm}, clip]{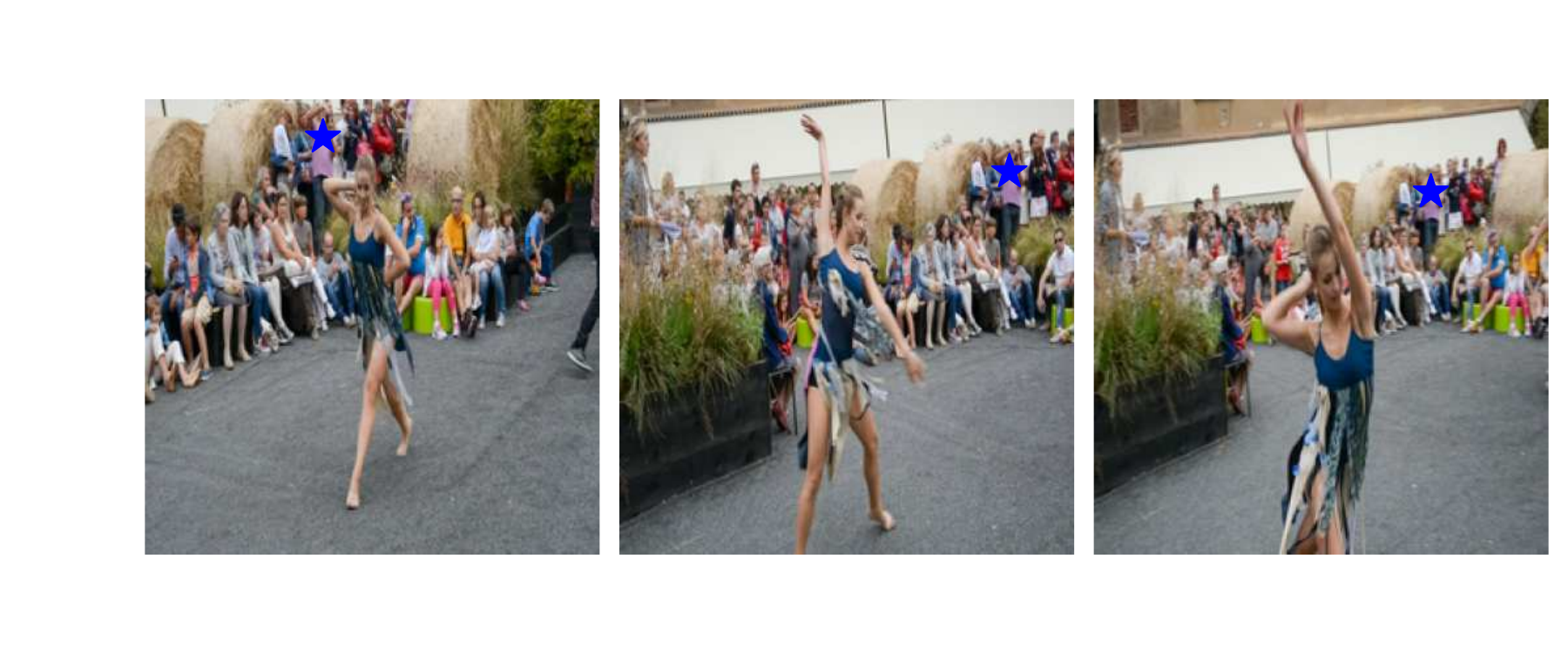}}\vspace{-0.1cm}
        \subfloat{\includegraphics[width=.99\linewidth, trim={0cm 1.5cm 0cm 2cm}, clip]{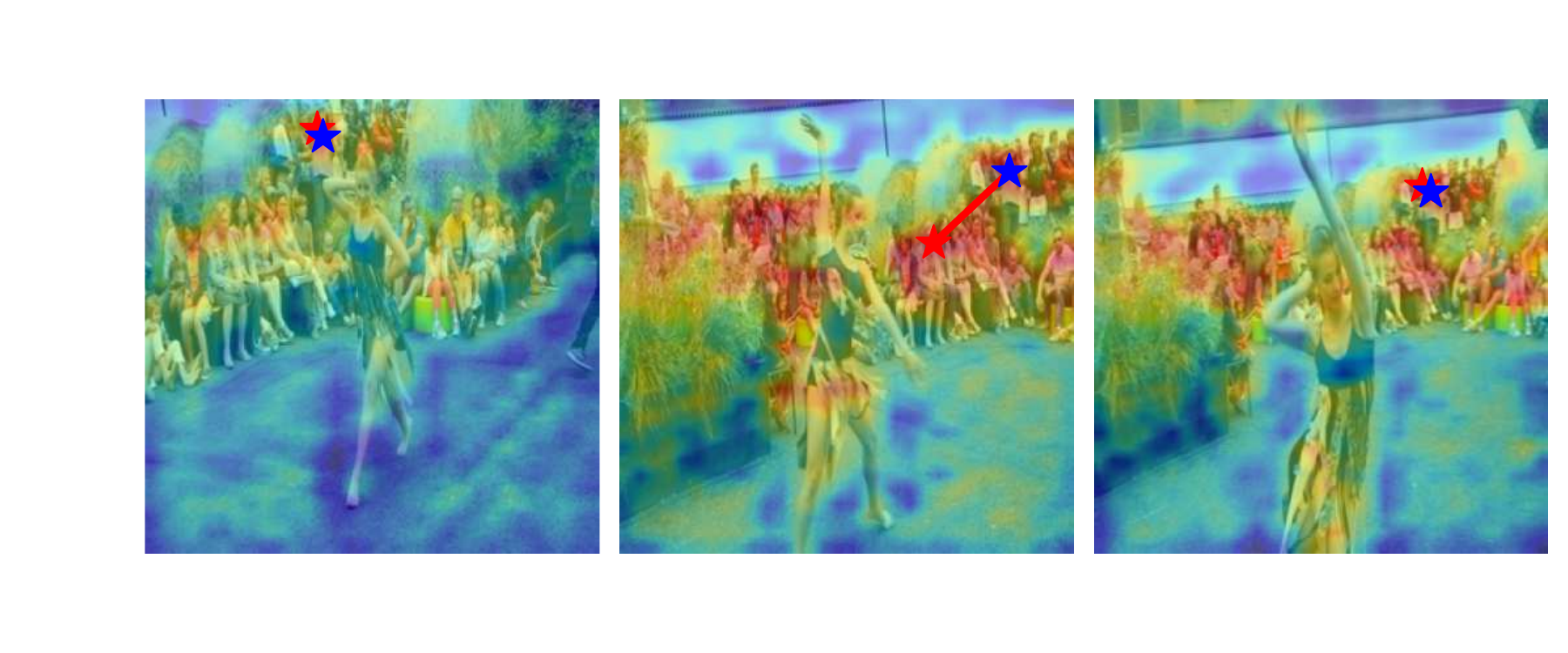}}\vspace{-0.1cm}
        \subfloat{\includegraphics[width=.99\linewidth, trim={0cm 1.5cm 0cm 2cm}, clip]{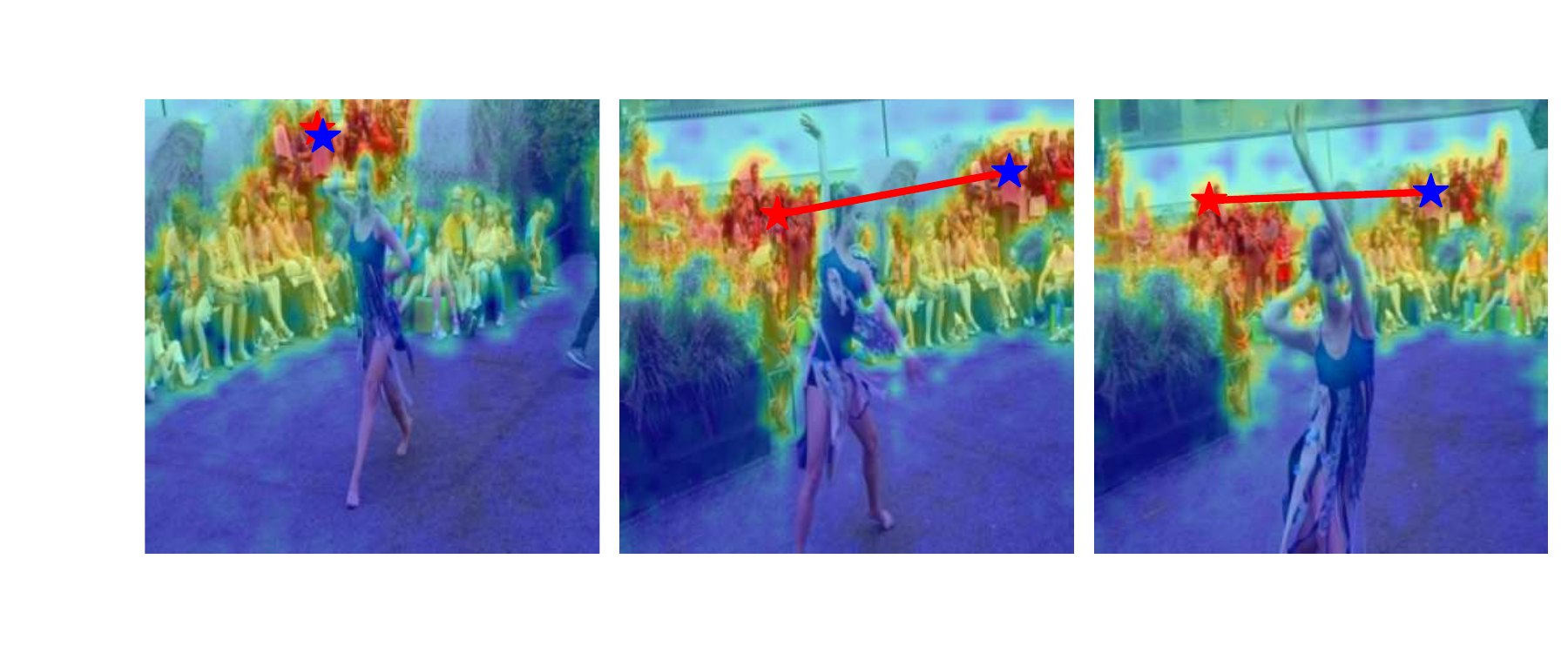}}  \vspace{-0.1cm}
        \subfloat{\includegraphics[width=.99\linewidth, trim={0cm 1.5cm 0cm 2cm}, clip]{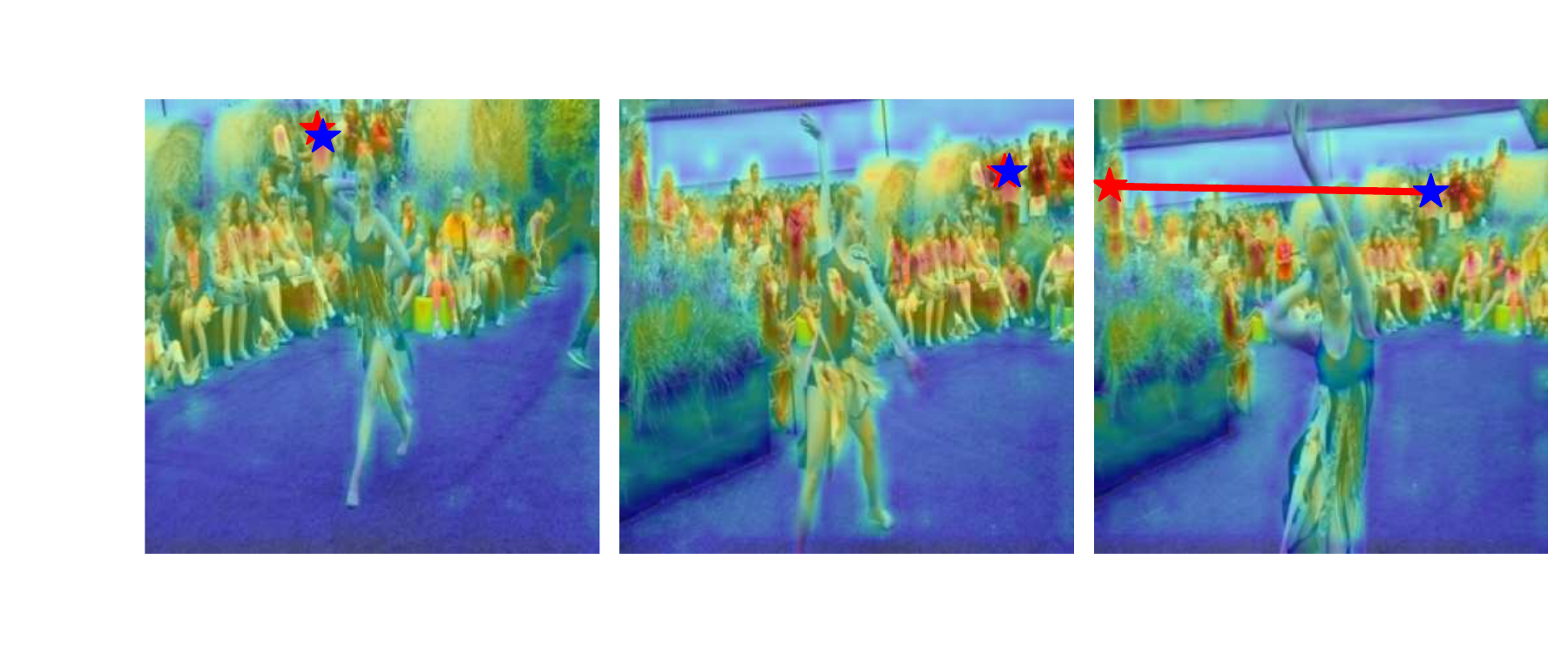}}
    \end{minipage}%
    \begin{minipage}{0.33\textwidth}
        \subfloat{\includegraphics[width=.99\linewidth, trim={0cm 1.5cm 0cm 2cm}, clip]{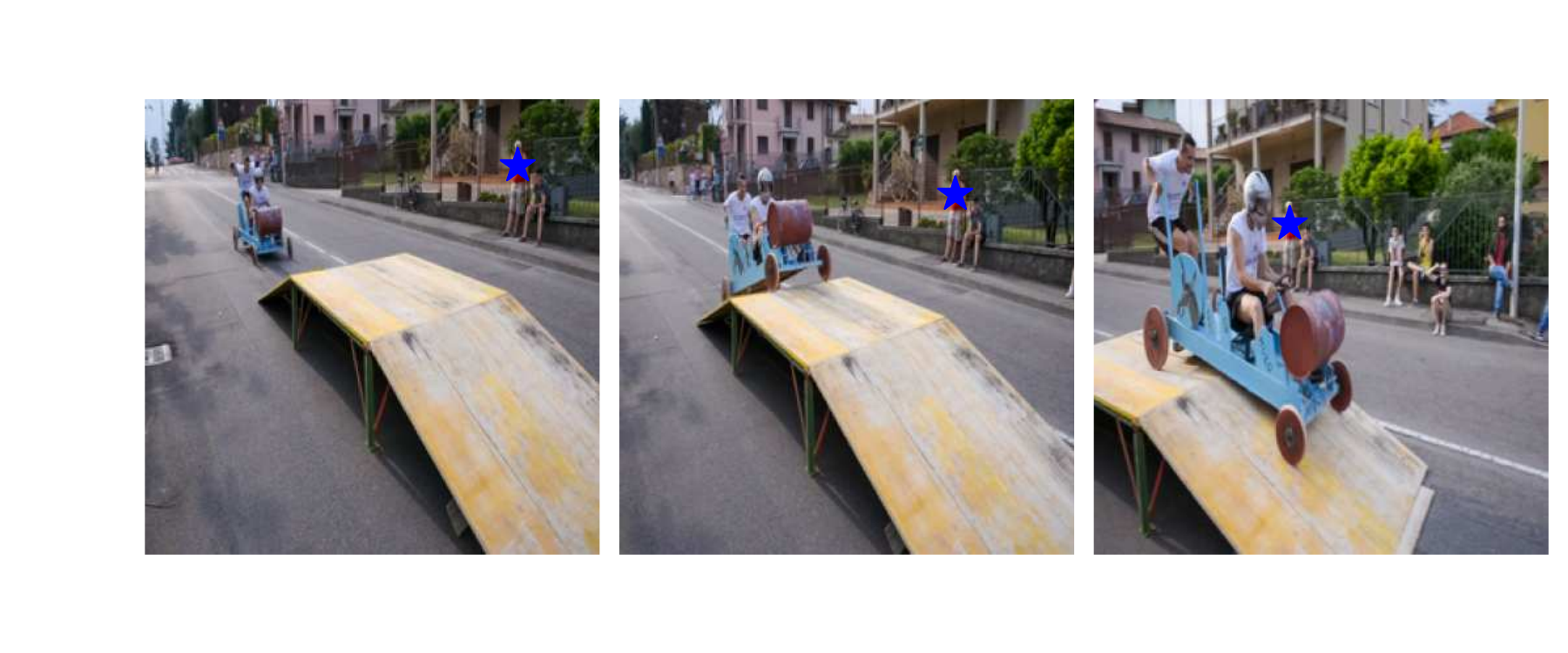}}\vspace{-0.1cm}
        \subfloat{\includegraphics[width=.99\linewidth, trim={0cm 1.5cm 0cm 2cm}, clip]{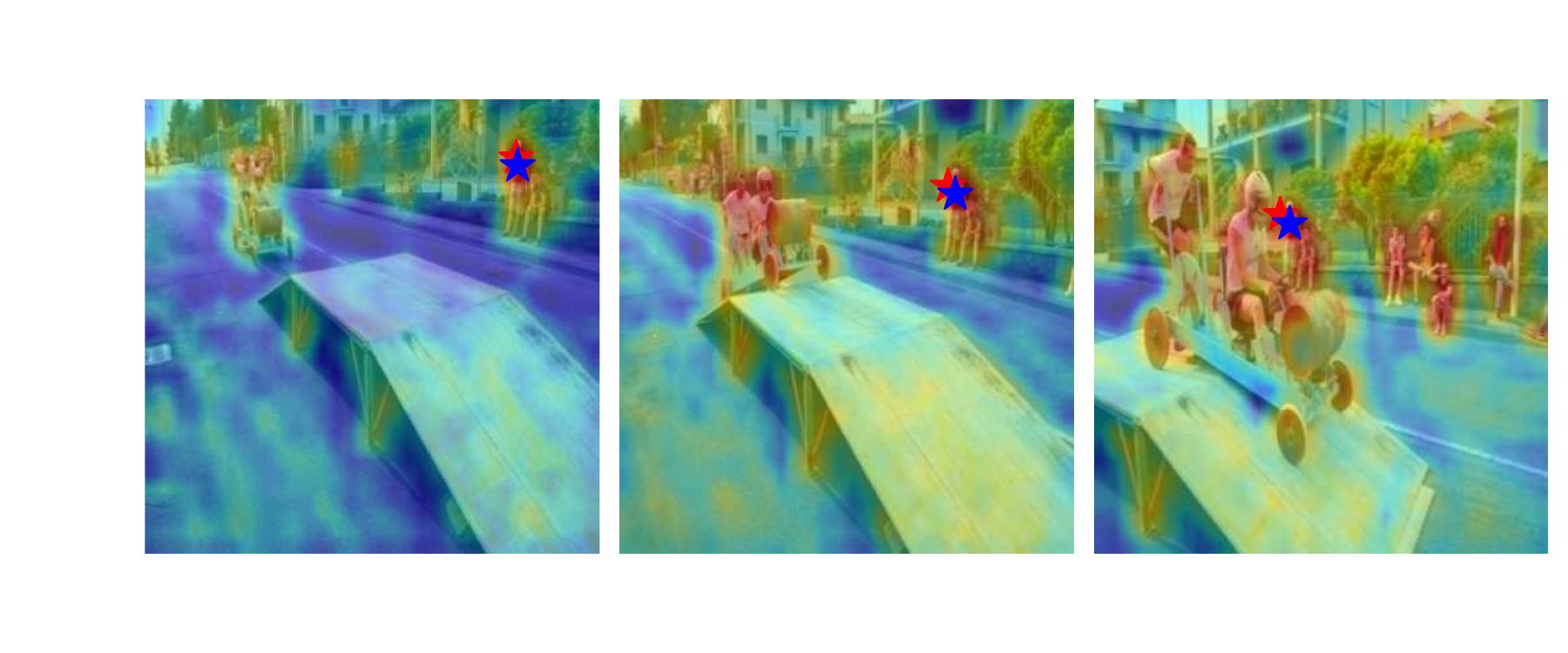}}\vspace{-0.1cm}
        \subfloat{\includegraphics[width=.99\linewidth, trim={0cm 1.5cm 0cm 2cm}, clip]{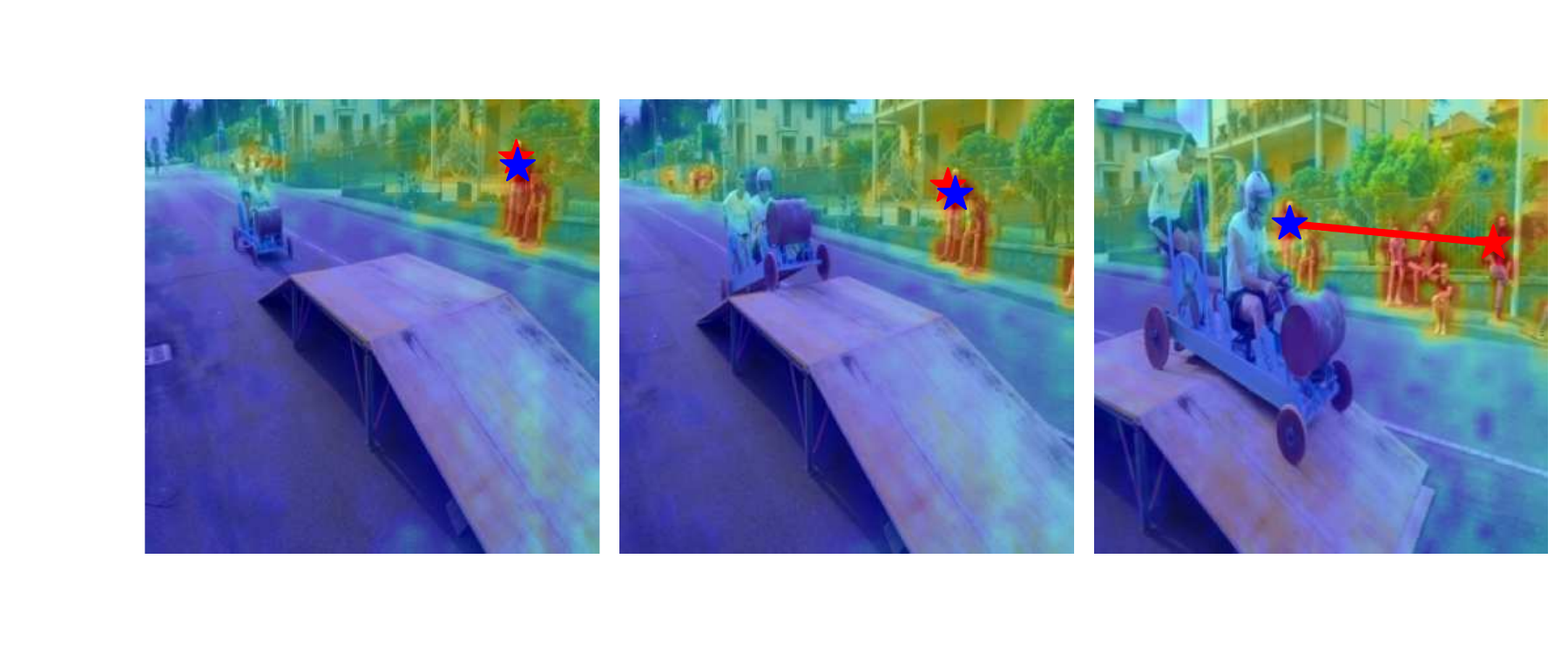}}  \vspace{-0.1cm}
        \subfloat{\includegraphics[width=.99\linewidth, trim={0cm 1.5cm 0cm 2cm}, clip]{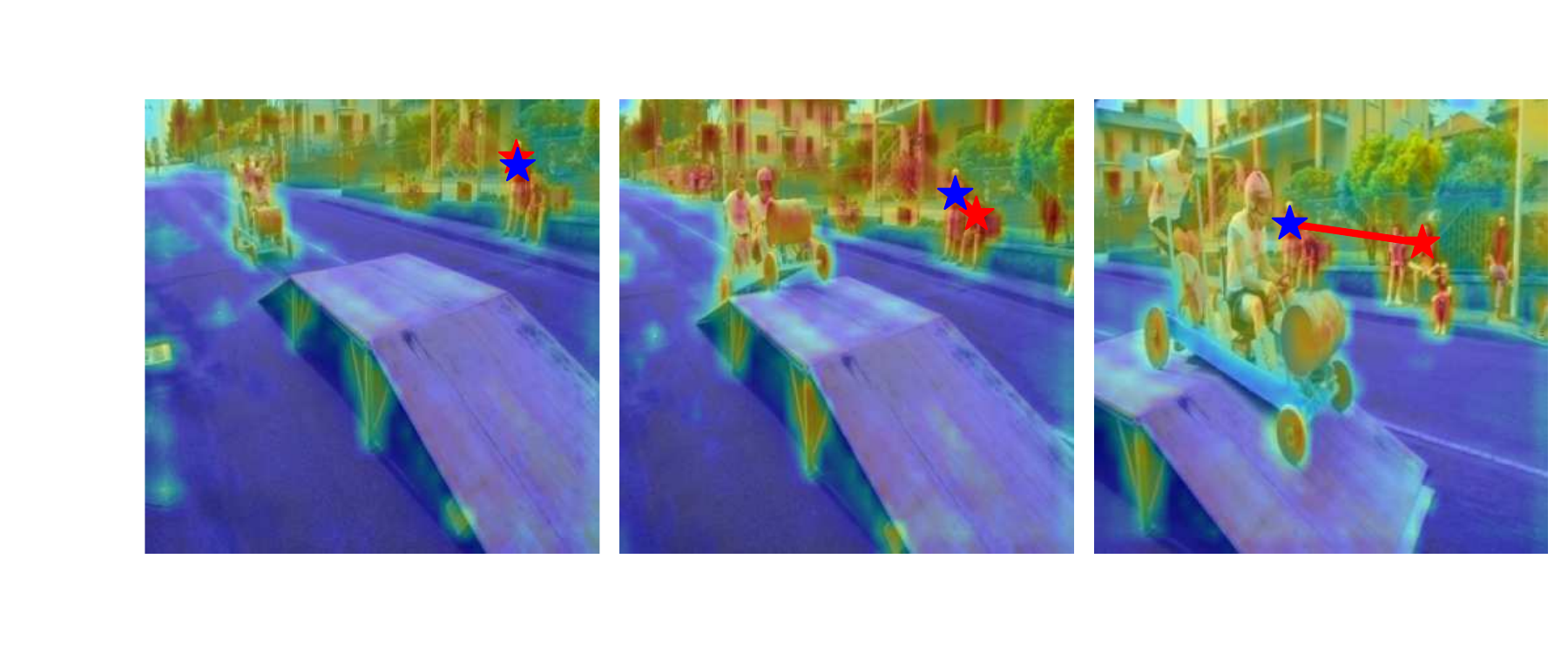}}
    \end{minipage}%

    \caption{\textbf{Qualitative Results of Zero-Shot Point Tracking on TAP-Vid DAVIS.} This figure shows the performance of zero-shot point tracking on TAP-Vid DAVIS. Query points from three videos are used to generate correlation map comparing the sampled query feature to features from different frames. Real correspondences are shown in the upper row. The models evaluated are Stable Diffusion~\cite{Rombach2022CVPR}, DINOv2~\cite{Oquab2023ARXIV}, and SAM~\cite{Kirillov2023ICCV}. Predictions are indicated by {\color{red} red stars}, representing the most similar locations, while ground truth correspondences are shown as {\color{blue} blue stars}. Red lines connect them to illustrate the spatial difference, or error, between the ground truth and the prediction. Warmer colors in the cost volumes indicate higher similarity.}
    \label{fig:qualitative}
\end{figure*}

%% file: tables/zero_shot_architecture.tex
\begin{table}[b]
    \centering
    \small
    \caption{\textbf{Effect of Architecture for Zero-Shot Evaluation.} This table shows the zero-shot evaluation results for different architectures of DINOv2 by varying the resolution on TAP-Vid DAVIS.}
    \begin{tabular}{l c | c}
        \toprule
        \textbf{Architecture} & \textbf{Final Resolution} & $\delta^\text{vis}_\text{avg}$  \\
        \midrule
        ViT-\textbf{S}/14 & 32 x 32 & 37.1 {\color{red} \scriptsize (-0.9)}  \\
        ViT-\textbf{B}/14 & 32 x 32 & 38.0 \\
        ViT-\textbf{L}/14 & 32 x 32 & 39.1 {\color{ForestGreen} \scriptsize (+1.1)} \\
        ViT-\textbf{G}/14 & 32 x 32 & 40.0 {\color{ForestGreen} \scriptsize (+2.0)}\\
        \midrule
        ViT-\textbf{B}/14 & 16 x 16 & 24.9 {\color{red} \scriptsize (-13.1)}\\
        ViT-\textbf{B}/14 & 48 x 48 & 45.1 {\color{ForestGreen} \scriptsize (+7.1)} \\
        ViT-\textbf{B}/14 & 64 x 64 & 48.5 {\color{ForestGreen} \scriptsize (+10.5)}\\
        
        \bottomrule
    \end{tabular}
    \label{tab:zero_shot_arch}
\end{table}

%% file: tables/probing.tex
\begin{table}[t]
    \centering
    \small
    \setlength{\tabcolsep}{3pt}
    \caption{\textbf{Probing and Adapting DINOv2.} This table shows different setups for DINOv2 ViT-S/14 on the TAP-Vid DAVIS, including the number of learnable parameters. The setups include zero-shot, probing, and adaptation with various LoRA ranks.}
    \begin{tabular}{l c c c | c c c}
        \toprule
        \textbf{Model} & \textbf{Setup} & \textbf{Rank} & \textbf{\#L.P.} & \text{AJ} & $\delta^\text{vis}_\text{avg}$   & \text{OA} \\
        \midrule
        DINOv2 & \text{Zero-shot} & - & 0 & - & 37.1 & - \\
        DINOv2 & \text{Probing} & - & 5.5K & 27.1 & 42.3 & 79.4 \\
        DINOv2 & \text{Adaptation} & 16 & 0.3M & 33.4 & 49.0 & 80.1 \\
        DINOv2 & \text{Adaptation} & 32 & 0.6M & 33.9 & 49.7 & \textbf{80.4} \\
        DINOv2 & \text{Adaptation} & 64 & 1.2M & \textbf{35.0} & \textbf{51.3} & 80.2 \\
        \midrule
        TAPNet~\cite{Doersch2022NeurIPS} & \text{Supervised} & - & 12.0M &  33.0 &  48.6 &  78.8 \\
        \bottomrule
    \end{tabular}
    \label{tab:probing}
\end{table}

%% file: sec/05-discussion.tex
\section{Discussion}
\label{sec:discussion}
\quad We explored the geometric awareness of vision foundation models for long-term point tracking under zero-shot settings and showed that Stable Diffusion has better geometric understanding than other models. Moreover, our experiments demonstrate that DINOv2 can surpass the performance of supervised models despite a significantly lighter training setup, indicating that these models inherently possess the notion of correspondence. However, our analysis is limited to leveraging the correlation maps in a single time step. Using features directly to better handle occlusions or integrating multi-frame designs~\cite{Karaev2024ECCV} to better manage feature drifts are promising directions for future research.